\documentclass[letterpaper, 10 pt, conference]{ieeeconf}  

\IEEEoverridecommandlockouts                              

\usepackage{cite}
\usepackage{amsmath,amssymb,amsfonts}
\usepackage{graphicx}
\usepackage{subcaption}
\usepackage{array}

\title{\LARGE \bf
Co-Design of Out-of-Distribution Detectors for Autonomous Emergency Braking Systems*
}

\author{Michael Yuhas$^{1,2}$ and Arvind Easwaran$^{2}$
\thanks{*This research was funded in part by MoE, Singapore, Tier-2 grant number MOE2019-T2-2-040.  This research is part of the programme DesCartes and is supported by the National Research Foundation, Prime Minister’s Office, Singapore under its Campus for Research Excellence and Technological Enterprise (CREATE) programme.}
\thanks{$^{1}$Energy Research Institute @ NTU, Interdisciplinary Graduate Programme, Nanyang Technological University, Singapore}%
\thanks{$^{2}$School of Computer Science and Engineering, Nanyang Technological University, 50 Nanyang Avenue, Singapore 639798
        {\tt\small \{michaelj004,arvinde\}@e.ntu.edu.sg}}%
}

\begin{document}

\maketitle
\thispagestyle{empty}
\pagestyle{empty}

\begin{abstract}

Learning enabled components (LECs), while critical for decision making in autonomous vehicles (AVs), are likely to make incorrect decisions when presented with samples outside of their training distributions.  Out-of-distribution (OOD) detectors have been proposed to detect such samples, thereby acting as a safety monitor, however, both OOD detectors and LECs require heavy utilization of embedded hardware typically found in AVs.  For both components, there is a tradeoff between non-functional and functional performance, and both impact a vehicle's safety.  For instance, giving an OOD detector a longer response time can increase its accuracy at the expense of the LEC.  We consider an LEC with binary output like an autonomous emergency braking system (AEBS) and use risk, the combination of severity and occurrence of a failure, to model the effect of both components' design parameters on each other's functional and non-functional performance, as well as their impact on system safety.  We formulate a co-design methodology that uses this risk model to find the design parameters for an OOD detector and LEC that decrease risk below that of the baseline system and demonstrate it on a vision based AEBS.  Using our methodology, we achieve a 42.3\% risk reduction while maintaining equivalent resource utilization.

\end{abstract}

\section{INTRODUCTION}
When learning enabled components (LECs) are exposed to data outside their training distribution, they are unlikely to make correct decisions.  In safety critical systems like autonomous vehicles, such a failure could lead to catastrophic results.  Out-of-distribution detectors have been proposed to detect such data~\cite{cai2020}, however, the introduction of an OOD detector exposes the system to additional risks.  If the OOD detector does not yield a decision before its deadline or returns a false negative result, it provides no protection.  Furthermore, when the OOD detector shares the same computational resource (like an embedded GPU) with an LEC, it interferes with the LEC's ability to meet deadlines~\cite{yuhas2022b}.  Additionally, false positives from the OOD detector can affect the system's availability, leading to a decrease in performance~\cite{alecu2022}.
\begin{figure}
    \centering
    \begin{subfigure}[b]{0.45\textwidth}
        \centering
        \includegraphics[width=1\linewidth]{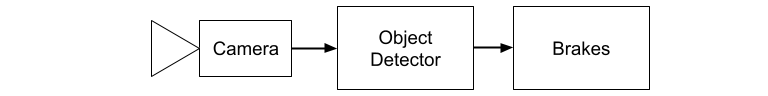}
        \caption{\footnotesize{Baseline AEBS; only the object detector triggers braking.}}
        \label{subfig:baseline}
    \end{subfigure}
    \begin{subfigure}[b]{0.45\textwidth}
        \centering
        \includegraphics[width=1\linewidth]{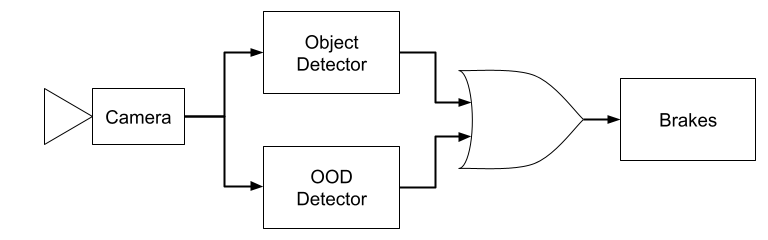}
        \caption{\footnotesize{OOD detector acts as a safety monitor for the object detector.}}
        \label{subfig:sys_new}
    \end{subfigure}
    \caption{\footnotesize{Block diagram of the AEBSs considered in this paper.}}
    \label{fig:block_diagram}
    \vspace{-5mm}
\end{figure}

Given these challenges, we seek to co-design an OOD detector and an LEC such that the new system (Fig.~\ref{subfig:sys_new}) uses the same hardware platform as the original design (Fig.~\ref{subfig:baseline}), but positively impacts safety.  As both the OOD detector and the LEC are implemented with deep neural networks (DNNs), we identify a subset of hyperparameters that can be selected independently for both components that tradeoff functional and non-functional performance.  However, this problem is different from a typical hyperparameter selection problem.  First, the objective is a function of both deadline misses and the functional performance of each network; it is not enough to minimize one of these values constrained on the others.  For example, a DNN with higher accuracy may be able to tolerate a greater level of deadline misses.  Second, the parameter selection for the each of the networks is not independent.  For example, choosing a parameter for the OOD detector that lengthens execution time and increases accuracy will affect the response time of the LEC.

We use risk as an objective to combine the functional and non-functional performance of both components and express the system risk for a generic binary classification task in terms of its components' failure probabilities.  We then apply this model to a simplified autonomous emergency braking system (AEBS), which functions as a binary classifier (emergency stop or no action) for a given input sample.  The contributions of this paper are as follows:
\begin{itemize}
    \item We derive a risk model valid for any binary classification task that expresses risk in terms of the co-design hyperparameters of an LEC and an OOD detector.
    \item We formulate a co-design methodology that efficiently explores the design space of OOD detector and LEC hyperparemeters to minimize risk without exceeding the average utilization of the baseline system.
    \item We co-design an OOD detector and object detector for a vision based AEBS to demonstrate OOD detectors' potential for risk reduction when deployed intelligently.  We show up to a 42.3\% risk reduction while maintaining the same average resource utilization as the baseline.
\end{itemize}

\section{BACKGROUND}
\subsection{Out-of-Distribution Detection}
Building functional OOD detectors has been well studied.  Detectors like ODIN~\cite{hsu2020} use the intermediate layers of a DNN to identify OOD samples.  The disadvantage is that the functional performance of the OOD detector is tightly coupled with the LEC it is monitoring.  In~\cite{an2015}, An and Cho proposed an OOD detector whose weights are trained independently and therefore conditionally independent of an LEC given a shared training set.  They used the reconstruction loss of a variational autoencoder (VAE) to determine whether or not a given sample was drawn from the training distribution.  In~\cite{cai2020}, Cai and Koutsoukos demonstrated the effectiveness of the reconstruction based OOD detector for autonomous driving and used inductive conformal prediction (ICP) and a martingale to deal with the time dimension.  However, reconstructing an image via VAE is costly in terms of execution time.  In~\cite{vasilev2019}, OOD detection in the latent space of a VAE was proposed, which only required running the encoder portion of the VAE.  Ramakrishna \textit{et al.} showed that the latent space of a VAE could be partially disentangled to detect OOD samples from different generative factors (e.g., rain, brightness, etc.), and demonstrated this on an autonomous driving dataset~\cite{ramakrishna2022}.  Although design methodologies have been proposed to optimize the execution time of such an OOD detector while respecting bounds on accuracy~\cite{yuhas2022a}, they do not take into account the scenario where an OOD detector and an LEC share the same set of computational resources.

\subsection{Co-Design frameworks for LECs}
Previous literature has focused on risk analysis and risk reduction as a means to safely deploy LECs in AVs. In~\cite{hartsell2021}, the ReSoNate framework was proposed, which calculated run time risk given environmental state and internal state and used this information to select the best controller for a particular scenario.  Although this framework used OOD detectors for safety monitoring, it did not account for the effect of an OOD detector on the LEC's response time.  In~\cite{bekiaris2005, becker2020, sun2021}, traditional risk analysis techniques were applied to automotive systems containing LECs, but no concrete methodology to apply them toward the co-design problem was demonstrated. In~\cite{alecu2022}, Alecu \textit{et al.} described the problem of LEC and monitor design as a tradeoff between safety and availability, however, they only explored this tradeoff in terms of functional performance, not response times.

Other works have focused on the deployment of multiple DNNs to shared resources while respecting schedulability.  In~\cite{kang2021}, an algorithm was presented for scheduling multiple DNNs across a CPU and GPU that took into account accuracy loss due to quantization.  While the deadlines of both tasks were respected, the algorithm could not actively trade accuracy between its scheduled tasks to minimize a shared risk objective.  In~\cite{ling2022}, DNNs were also divided across resources and their accuracy bounds were checked before scheduling.  A neural architecture search could improve the accuracy of a DNN if response time allowed, but it could not take into account the cumulative effect of multiple DNNs.

\section{PROBLEM DEFINITION}
\label{sec:probdef}

\subsection{Risk Minimization over Design Parameters}
We consider a system with two DNNs like the AEBS in Fig.~\ref{subfig:sys_new}.  We refer to the object detector as the \textit{essential component} (EC), and provide a generalized risk model for any EC that implements a binary classification task.  Let the EC be defined as $f(x;\theta_{EC},\lambda_{EC},\tau_{EC}):\mathbb{R}^{n}\mapsto \{ 0,1 \}$.  Here, $x$ is an $n$ dimensional input sample, e.g., radar point cloud or image.  The EC maps this to one of two values: a $0$ indicates a negative result and a $1$ indicates a positive result. $f$ is parameterized by $\lambda_{EC}$, a set of hyperparemeters determining the structure of $f$, e.g., number of layers or input image size; $\theta_{EC}$, a set of parameters learned during training; and $\tau_{EC}$, the confidence threshold at which the EC returns a $1$.  Likewise, let the OOD detector be defined as $g(x;\theta_{OOD},\lambda_{OOD},\tau_{OOD}):\mathbb{R}^{n}\mapsto \{ 0, 1 \}$.  Once again $x$ is an $n$ dimensional input data sample that the OOD detector maps to $0$ for in-distribution (ID) or $1$ for OOD.  The OOD detector is parameterized by $\lambda_{OOD}$, a set of hyperparameters determining the structure of $g$; $\theta_{OOD}$, the parameters learned during training; and $\tau_{OOD}$, the confidence threshold at which the OOD detector returns a $1$. Let $\Lambda = (\lambda_{EC}, \lambda_{OOD})$ be a tuple of the hyperparameters that affect the response times and functional performance of both components, while $T = (\tau_{EC}, \tau_{OOD})$ and $\Theta = (\theta_{EC}, \theta_{OOD})$ affect functional performance only.

It is impossible to design an OOD detector or EC with zero failure rate. We are interested in the negative effects that occur if these components fail, how likely failures are to occur, and how severe their consequences would be.  Risk, a combination of the severity of an event with its probability is a natural way to measure this, and is found in automotive safety standards like ISO 26262~\cite{salay2017}.  We define risk mathematically in~(\ref{eq:risk}), where $\mathcal{R}$ is the system's total risk, $\mathcal{E}$ is the set of all hazardous events that can occur, $P(x)$ denotes the probability of event $x$, and $S(x)$ denotes its severity.
\begin{equation}
\label{eq:risk}
    \mathcal{R}=\sum_{E_i\in\mathcal{E}}{P(E_i)S(E_i)}
\end{equation}
Our goal is to design a system that minimizes risk as defined in~(\ref{eq:risk}), such that the average resource utilization does not exceed that of the baseline as shown in~(\ref{eq:minimization}).
\begin{align*}
    \underset{\Lambda,T,\Theta}{\operatorname{argmin}} & {\sum_{E_i\in\mathcal{E}}{P(E_i|\Lambda,T,\Theta})S(E_i)}\\
    \text{s.t.}\quad &\bar{U}(\Lambda) \le \bar{U}_{base}\addtocounter{equation}{1}\tag{\theequation} \label{eq:minimization}
\end{align*}
Here, $\bar{U}(\Lambda)$ is the average resource utilization of a system given structural parameters $\Lambda$ and $\bar{U}_{base}$ is the average resource utilization of the baseline system (e.g., Fig.~\ref{subfig:baseline}).  We define $\bar{U}$, average resource utilization, as the percentage of time the shared computational resource is occupied in one period (from both jobs' release times to their deadline) averaged over all periods. Constraining utilization is reasonable because the EC plus OOD detector should be a drop-in replacement for the baseline system; if the baseline utilization is exceeded, it may interfere with other system-level components.  Note that by changing the design parameters $\Lambda$, $T$, and $\Theta$, we can affect the probability of an event occurring, but we cannot change its severity. While the learned parameters $\Theta$ affect the probability of an event occurring, minimizing risk subject to these learned parameters is outside the scope of this work as it can be optimized through existing training methods like~\cite{reddi2018}.
\begin{table}
  \vspace{1mm}
  \caption{\textsc{\footnotesize{Assumptions under which our risk model holds.}}}
  \label{tab:assumptions}
  \begin{tabular}{|>{\centering\arraybackslash}p{0.075\linewidth}|p{0.8\linewidth}|}
    \hline
    \textbf{Asm.}&\textbf{Description}\\
    \hline
    A1 & Negligible risk for correct classification \\
    A2 & No reject option \\
    A3 & Hazard avoidance when EC \textit{or} OOD detector returns $1$ \\
    A4 & Results of both components are independent across time\\
    &(\emph{i.e.,} dependency across successive inputs is ignored)\\
    A5 & If EC or OOD detector does not complete before its deadline, all work is discarded and execution is terminated\\
    A6 & No action is triggered due to early termination \\
    A7 & OOD detector missing a deadline is independent of its detection result\\
    A8 & Functional results of the EC and OOD detector are conditionally independent given $\lnot E_e$ or $\lnot E_\epsilon$\\
    \hline
\end{tabular}
\vspace{-5mm}
\end{table}
In order to solve this minimization problem, we need to identify the events that compose $\mathcal{E}$.  Our assumptions are listed in Table~\ref{tab:assumptions}.  Under A1, the risk induced by the EC making a correct decision (true positive or true negative) is negligible and has already been minimized under safety of the intended functionality~\cite{chu2023}.  Under A2, the system must generate one of two outputs: $0$ or $1$; there is no reject option. This assumption is valid as even if the OOD detector correctly identifies an OOD input, the system must perform some action, even if that action is to keep doing the same thing while waiting to reprocess the rejected sample with another model~\cite{alecu2022}.  Under A3, we assume that either a binary classification result of $1$ \textit{or} an OOD detection result of $1$ will trigger a hazard avoidance action.  This reflects the use of OOD detectors in prior works~\cite{cai2020, hartsell2021, yuhas2022b} and leads to two possible failure modes, $\mathcal{E} = \{ E_0, E_1 \}$, where:
\begin{itemize}
    \item $E_0$ -- The event where no action is taken when a hazard is present (OOD detector and EC return $0$ while the ground truth for the EC is $1$)
    \item $E_1$ -- The event where corrective action is taken when no hazard is present (OOD detector returns $1$ \textit{or} EC returns $1$ while the ground truth for the EC is $0$)
\end{itemize}

\subsection{Determining Probabilities through Fault Tree Analysis}
\label{sec:risk}
We use fault tree analysis (FTA), a deductive method that works backwards from a given failure mode and determines which intermediate failures must occur to cause it.  FTA allows us to express the probabilities of the top-level events in terms of the probabilities of their generating events~\cite{lee1985}.  We want to express the probabilities of $E_0$ and $E_1$ with respect to the design parameters $\Lambda$, $T$, and $\Theta$.  Under A4, we assume the independence of detection results on different samples over time.  Although not strictly true in practice as observations in one time instance depend on the control action from the previous instance, we are dealing with feed forward DNNs, so previous results are not used directly in the computation of the next result.  Other works have made this assumption as well~\cite{abdelzaher2023}.

First, we define the intermediate events that could lead to the occurrence of $E_0$ and $E_1$.  Each execution of the EC can be considered as an experiment where the outcome is one of the events in Table~\ref{tab:yolo_events}. Note that $E_a$, $E_b$, $E_c$, $E_d$, and $E_e$ are mutually exclusive and $P(E_a)+P(E_b)+P(E_c)+P(E_d)+P(E_e)=1$.
\begin{table}
  \vspace{1mm}
  \caption{\footnotesize{\textsc{Sample space for the outcome of the EC.}}}
  \label{tab:yolo_events}
  \begin{tabular}{|>{\centering\arraybackslash}p{0.08\linewidth}|p{0.8\linewidth}|}
    \hline
    \textbf{Event}&\textbf{Definition}\\
    \hline
    $E_a$ & The event that the EC gives a false positive result \\
    $E_b$ & The event that the EC gives a true positive result \\
    $E_c$ & The event that the EC gives a false negative result \\ 
    $E_d$ & The event that the EC gives true negative result \\
    $E_e$ & The event that the EC misses its deadline\\
    \hline
\end{tabular}
\vspace{-2mm}
\end{table}
Likewise, Table~\ref{tab:ood_events} shows the events representing the possible outcomes of the OOD detector.  Similar to the previous case, $E_\alpha$, $E_\beta$, $E_\gamma$, $E_\delta$, and $E_\epsilon$ are mutually exclusive and $P(E_\alpha)+P(E_\beta)+P(E_\gamma)+P(E_\delta)+P(E_\epsilon)=1$.
\begin{table}
  \caption{\footnotesize{\textsc{Sample space for the outcome of the OOD detector.}}}
  \label{tab:ood_events}
  \begin{tabular}{|>{\centering\arraybackslash}p{0.08\linewidth}|p{0.8\linewidth}|}
    \hline
    \textbf{Event}&\textbf{Definition}\\
    \hline
    $E_\alpha$ & The event that the OOD detector gives a false positive result \\
    $E_\beta$ & The event that the OOD detector gives a true positive result \\
    $E_\gamma$ & The event that the OOD detector gives a false negative result \\ 
    $E_\delta$ & The event that the OOD detector gives true negative result \\
    $E_\epsilon$ & The event that the OOD detector misses its deadline \\
    \hline
\end{tabular}
\vspace{-5mm}
\end{table}
Under A5, we assume that as soon as any component misses its deadline for a sample $x$, all work is discarded and the execution is terminated; this assumption is also considered in other literature~\cite{pazzaglia2019}.  Under this assumption, while the OOD detector and EC missing a deadline for the same input are dependent events ($E_e \not\perp E_\epsilon$), these events are independent across samples.  Under A6, we assume that missing a deadline will not cause any corrective action. This is a common strategy in controls literature~\cite{vreman2022} and we use it here to reduce false positives.  Let $E^{base}_0$ and $E^{base}_1$ correspond to $E_0$ and $E_1$ in the baseline system.  The fault tree analysis trivially yields~(\ref{eq:ebase0}) and~(\ref{eq:ebase1}).
\begin{align}
    E^{base}_0 &= E_c \lor E_e \implies P(E^{base}_0) = P(E_c) + P(E_e)\label{eq:ebase0}\\
    E^{base}_1 &= E_a \implies P(E^{base}_1) = P(E_a)\label{eq:ebase1}
\end{align}
For the system with OOD detector, let $E^{mod}_0$ and $E^{mod}_1$ correspond to $E_0$ and $E_1$:
\begin{align}
    E^{mod}_0 &= \{E_\gamma \lor E_\delta \lor E_\epsilon\} \land \{E_c \lor E_e \land E_{pos}\}\\
    E^{mod}_1 &= \{ \{E_\beta \lor E_\alpha\} \land \lnot E_{pos} \} \lor E_a
\end{align}
Here, $E_{pos}$ is the event that a sample's ground truth is $1$.  Note that the OOD detector reduces the chance of $E^{mod}_0$ as a functional failure or deadline miss of both components is required.  However, the OOD detector increases the chance of $E^{mod}_1$, as any positive result leads to this failure as long as the ground truth is $0$.  Under A7, we assume that the OOD detector's probability of missing a deadline is not a function of the sample's ground truth.  This assumption is based on the architecture of the OOD detectors we consider~\cite{ramakrishna2022}.  We do not make the same assumption for the EC, as some architectures may take different amounts of time to reach a decision depending on the input data.  For example, in our AEBS use case, a YOLO object detector requires non-max suppression to select the best bounding boxes, so its execution time is dependent on the number of boxes identified~\cite{wang2022}.  Although the results of the OOD detector and the EC are dependent due to response time interference, under A8, we assume that given no deadline miss occurs, the functional results of both components are independent.  We make this assumption due to the use of an independently trained OOD detector like~\cite{an2015}.
\begin{align*}
    P(E_x|E_y) = P(E_x|\lnot E_e) \forall x \in \{\alpha, \beta, \gamma, \delta\}; y \in \{a, b, c, d\} \\
    P(E_x|E_y) = P(E_x|\lnot E_\epsilon) \forall x \in \{a, b, c, d\}; y \in \{\alpha, \beta, \gamma, \delta\}\addtocounter{equation}{1}\tag{\theequation} \label{eq:a37}
\end{align*}
The equalities in~(\ref{eq:a37}) arise from A8 and we use them to simplify $P(E^{mod}_0)$~(\ref{eq:e0}) and $P(E^{mod}_1)$~(\ref{eq:e1}).
\begin{align*}
    P(E^{mod}_0) &= P(Ec)(P(E_\gamma) + P(E_\delta) + P(E_\epsilon) \\
    & - P(E_\gamma|E_e) - P(E_\delta|E_e) - P(E_\epsilon|E_e)) \\
    & + P(E_{pos})P(E_e|E_{pos})(P(E_\gamma|E_e) \\
    & + P(E_\delta|E_e) + P(E_\epsilon|E_e))\addtocounter{equation}{1}\tag{\theequation} \label{eq:e0}
\end{align*}
$P(E^{mod}_0)$ is composed of two terms: the first takes into account the false negative rate of the EC and the probability that the OOD detector fails to override it while the second considers the case when the EC misses its deadline and the OOD detector fails to override.  Notice that an estimate of $P(E_{pos})$ is required to calculate this probability.
\begin{align*}
    P(E^{mod}_1) &= P(E_a) + P(E_\alpha)(1 - P(E_{pos})) \\
    & + P(E_\beta)(1 - P(E_{pos})) - P(E_a)(P(E_\alpha) \\
    & + P(E_\beta) - P(E_\alpha|E_e) - P(E_\beta|E_e))\addtocounter{equation}{1}\tag{\theequation} \label{eq:e1}
\end{align*}
$P(E^{mod}_1)$ is composed of three terms.  Either a false positive from the EC, or a positive result from the OOD detector when the ground truth is $0$ (terms one and two) can trigger this event.  The subtraction of term three removes the union of these two events as they are not mutually exclusive.

\section{CO-DESIGN METHODOLOGY}
The constituent terms of~(\ref{eq:e0}) and~(\ref{eq:e1}) can be estimated empirically for an EC and OOD detector pair with unique set of parameters $\Lambda$, $T$, and $\Theta$.  However, evaluating the risk for one pair involves training two separate DNNs, which means exploring the design space of all parameter combinations is prohibitively expensive, so we propose a design methodology to minimize the training time needed to find a satisfactory solution.
\begin{figure}
    \vspace{1.5mm}
    \centering
    \includegraphics[width=1\linewidth]{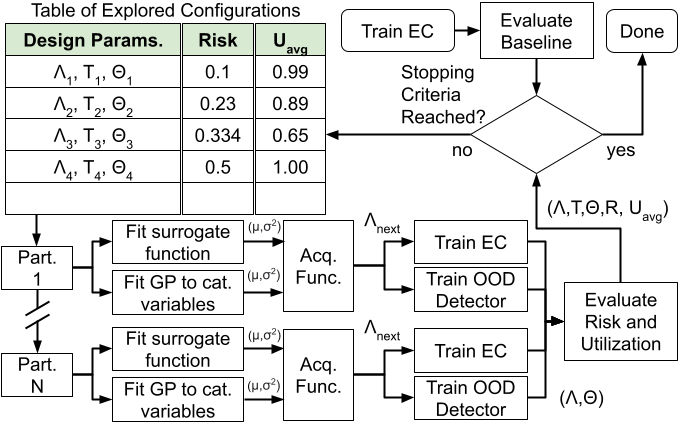}
    \caption{\footnotesize{Risk-aware co-design methodology for OOD detector and EC based on Bayesian optimization.}}
    \label{fig:framework}
    \vspace{-5mm}
\end{figure}
We note the following facts about our minimization problem, which inform our design strategy.  Firstly, for fixed hyperparameters $\Lambda$ and $T$, the resulting probability of an event $P(E_i)$ is not fixed. This is because training a DNN is a stochastic process, and retraining will result in a different set of learned parameters $\Theta$.  Next, we note that $\Lambda$ can be comprised of both numerical and categorical variables. An example of a numerical variable is the scaled input size to the EC or the OOD detector, while an example of a categorical variable is the architecture used for a particular component (e.g., a $\beta$-VAE OOD detector or a reconstruction based OOD detector). The presence of categorical variables means that gradient based optimization techniques are not an option.  While it is desirable to evaluate as few values of $\Lambda$ and $\Theta$ as necessary, once we have trained two DNNs for a $(\Lambda,\Theta)$ pair, it is relatively easy to find the $T$ that minimizes risk for that configuration. This is because during testing, the EC and OOD detector will produce a list of confidence scores for a given dataset and the functional performance at multiple thresholds can be computed without the need for the DNNs to re-infer the entire training set.

Since we are dealing with a possibly non-convex, noisy risk function, we propose a modified version of Bayesian optimization~\cite{malu2021} to find the parameters $\Lambda$ and $T$, that minimize risk.  A visualization of our design methodology is shown in Fig.~\ref{fig:framework}.  A table keeps track of all the design parameter combinations that have been tried and their respective risk and average utilization values.  First, the EC for the baseline system with no OOD detector is trained and the risk and average utilization are evaluated; the average utilization will serve as a constraint when evaluating proposed solutions.  Next, the search space is divided into $N$ partitions of equal size.  By dividing the search space into partitions, we can help reduce the likelihood that the Bayesian optimization gets stuck at a local minimum~\cite{wang2020}.  To start, we populate the table with $n_{init}$ entries with $\Lambda$s uniformly sampled across each partition.  Given the solutions already evaluated in each partition, we fit a \textit{surrogate function}.  The surrogate function estimates of the true risk value at any point as well as the confidence in that estimate.  We accomplish this with one Gaussian process model for all numerical parameters in $\Lambda$~\cite{hu2022} and use a separate Gaussian process model for each categorical variable if present~\cite{nguyen2020}.  Next, in each partition an \textit{acquisition function} uses the mean and variance from each surrogate function to generate a numerical estimate for how beneficial it would be to evaluate a new sample at a given point.  The choice of acquisition function affects the balance between exploiting existing good solutions and exploring other areas in the risk function's domain; we choose \textit{expected improvement}~\cite{hu2022}.  The $\Lambda$ that maximizes the acquisition function is determined numerically using the conjugate gradient method and then used to train a new EC and OOD detector.  Risk and utilization are experimentally evaluated for the $(\Lambda,\Theta)$ pair in each partition and $T$ is swept across the its entire range for the EC and OOD detector.  The solution at the $T$ that minimizes risk is compared with the baseline risk and average utilization.  If the utilization constraint is satisfied, the tuple $(\Lambda, T, \Theta, \mathcal{R}, \bar{U})$ is recorded in the table of previously evaluated points.  However, if the constraint is violated, its risk is set to an arbitrarily high value and then recorded in the table to encourage the acquisition function to look elsewhere.  We choose lack of improvement in risk after a set number of iterations as the stopping criteria.  While Bayesian optimization is not guaranteed to converge or find the minimum solution, it provides a powerful tool to deal with such a design problem.

\section{CASE STUDY: YOLO BASED AEBS}
In order to show the effectiveness of the proposed risk minimization strategy, we conduct a case study on a YOLO (You Only Look Once) based AEBS with OOD detector. This AEBS uses monocular vision to identify obstacles as demonstrated in~\cite{wu2022}. The output of the YOLO object detector is processed to make a binary decision: $0$ (do nothing) or $1$ (engage emergency braking).  We use images with heavy rain (not present during YOLO or OOD detector training) as OOD samples.  The code to generate our dataset, train the LECs, and execute the tests is publicly available\footnote{https://github.com/CPS-research-group/CPS-NTU-Public/tree/ITSC2023}.

\subsection{Dataset}
We use CARLA simulator version 0.9.13~\cite{dosovitskiy2017} to simulate an autonomous vehicle in an urban environment. $100$ video clips of $500$ frames at $30$ FPS are captured in CARLA built-in town $10$ (urban environment).  The ego vehicle uses the default autopilot to navigate the streets and additional vehicles and pedestrians are spawned into the town to add obstacles on the road.  The ego vehicle is selected as the Audi E-Tron with a forward facing RGB camera and segmentation camera affixed at relative coordinates $(x=10.5,y=0,z=0.7)$. The outputs of both cameras are resized to $800\times600$ pixels. The default time of day and weather are used for all the gathered clips.
\begin{figure*}
    \vspace{2mm}
    \centering
    \includegraphics[width=1\textwidth]{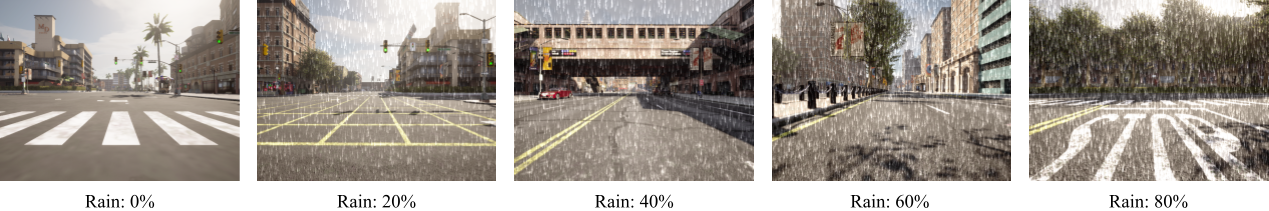}
    \caption{\footnotesize{Example images for selected rain levels in our dataset.}}
    \label{fig:dataset}
    \vspace{-5mm}
\end{figure*}
We use the same method as~\cite{yuhas2022a} to augment the images with varying amounts of rain. Image augmentation is applied such that $10$ rain levels ranging from ${0, 0.1, 0.2, \dots 0.9}$ are applied to $10$ different clips each containing $500$ images, resulting in $5000$ total images at each rain level in the dataset ($5\times10^4$ images in total).  Some examples of the generated images are shown in Fig.~\ref{fig:dataset}.

\subsection{Object Detector}
As proposed in~\cite{wu2022}, we use YOLO to perform object detection. Specifically, we select the YOLOv7 tiny model~\cite{wang2022}, as this provides state of the art object detection performance while keeping execution time low with respect to other YOLO variants.  Our free parameter $\lambda_{EC}$ for the co-design is the input image size, and the confidence threshold $\tau_{EC}$ is used by YOLO to determine if a bounding box is detected as an object or not.  We select $300$ images from clips $0-19$ (no rain and $10\%$ rain) as the training set ($6000$ images in total) and $100$ images from clips $0-19$ as the cross-validation set ($2000$ images in total). The ground truth bounding boxes were automatically obtained from the segmentation camera images by finding the contours of each region and the minimum bounding box that contained them.  Only the pedestrian and vehicle object classes were considered.  For this study images with $10\%$ rain or less are considered ID and images with more than $10\%$ rain are considered OOD. The $10\%$ cutoff was determined empirically as the point at which YOLO's performance begins to degrade.

To obtain the performance of the object detector in terms of $P(E_a)$, $P(E_b)$, $P(E_c)$, $P(E_d)$, and $P(E_e)$ we define two metrics unique to a vision based object detector. While the traditional false positive rate (FPR) of an object detector gives the rate of false positive bounding boxes per input image~\cite{padilla2021}, we are only interested in the case where at least one false positive is found, because this is enough to trigger emergency braking. Furthermore, even if a false positive detection occurs in a sample with another object present, it does not lead to the event $E_a$, since the vehicle is supposed to stop anyway. Our modified FPR ($FP_{m}$) metric is defined in~(\ref{eq:mfpr}), where $\mathcal{D}$ is a dataset consisting of tuples $(x,y)$; $x$ is an input image and $y$ is a list of objects present in the input. $f(x)$ is the trained object detector.
\begin{equation}
\label{eq:mfpr}
    FP_m(f,\mathcal{D}) = \frac{|\{(x,y)|f(x)=1 \land |y|=0; \forall (x,y) \in \mathcal{D}\}|}{|\mathcal{D}|}
\end{equation}
Likewise, we redefine false negative rate (FNR) as the number of inputs that contain an object, but where no bounding box is identified. Even if the object detector identifies the wrong object, this event will not contribute to $E_1$. Our modified FNR ($FN_{m}$) metric is defined in~(\ref{eq:mfnr}).
\begin{equation}
\label{eq:mfnr}
    FN_{m}(f,\mathcal{D}) = \frac{|\{(x,y)|f(x)=0 \land |y|>0; \forall (x,y) \in \mathcal{D}\}|}{|\mathcal{D}|}
\end{equation}

\subsection{OOD Detector}
For the OOD detector, we consider two free parameters that make up $\lambda_{OOD}$: the size of the input image and the OOD detector type ($\beta$-VAE or reconstruction based). The $\beta$-VAE OOD detector~\cite{ramakrishna2022} is designed to identify OOD samples caused by a specific generative factor. We select $200$ images from clips $0-19$ (no rain and $10\%$ rain) as the proper training set ($4000$ images in total), and $100$ images from clips $0-19$ for the calibration set ($2000$ images in total). This splitting strategy leaves us with the 2:1 train:calibration split recommended in~\cite{ramakrishna2022}.  The VAE portion of the OOD detector was constructed with four convolutional layers with depths 32/64/128/256 and a convolutional kernel of size $3$, each followed by a maxpool with kernel size $2$. The fully connected layers were sized $2048$, $1000$, $250$, and finally $50$ latent variables. All layers used a leaky ReLU activation function with $\alpha=0.1$ except for the final layer which used an identity activation function. The decoder was constructed as a mirror image of the encoder.  All variants of the network were trained for $350$ epochs using the Adam optimizer~\cite{reddi2018} with maximum learning rate set to $10^{-5}$. After training, the calibration set was used to select the latent variable that responded best to rain for each model, and the Kullback-Leibler divergence for each sample in the calibration set was used as a non-conformity score to train the detector's ICP.  For the reconstruction based OOD detector from~\cite{cai2020}, we used the same train:calibration split, but the OOD score is now the mean squared reconstruction error at the output of the decoder.  After training, the non-conformity scores between each sample in the calibration set and all the samples in the proper training set were calculated using the k-nearest neighbors algorithm with $k=|\mathcal{D}_{train}|$, the size of the training set.  These non-conformity scores were then used for ICP at the detector's output.

\subsection{Execution Dispatch and Timing}
To determine $P(E_e)$ and $P(E_\epsilon)$, we test the entire system on an Nvidia Jetson Nano with $2$ GB RAM and L4T 32.1 with the PREEMPT\_RT kernel patch installed. Both the YOLO object detector and OOD detector were executed on the embedded GPU, however, because the Jetson platform does not support NVIDIA MPS~\cite{NvidiaMPS}, both the detectors have to be submitted to the GPU sequentially, with the resource blocked until one is finished.  The distributions $P(E_e)$ and $P(E_\epsilon)$ are generated empirically by analyzing the response times on $1000$ images.  To calculate average utilization we measure the execution times of each job in a period and take the average percentage of time spent working on the two jobs across all periods. 

\subsection{Optimal Baseline AEBS}
\begin{figure}
    \centering
    \includegraphics[width=0.475\textwidth]{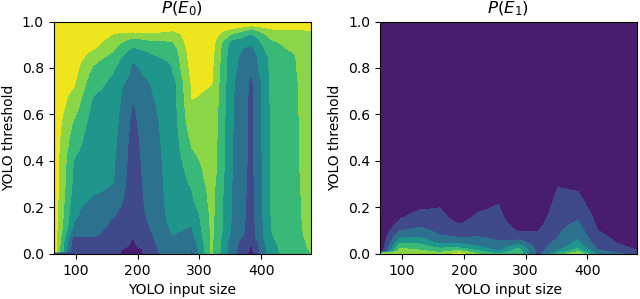}
    \caption{\footnotesize{$P(E^{base}_0)$, $P(E^{base}_1)$ across the entire design space of the baseline AEBS. Here, the deadline is set at $250$ ms ($4$ Hz). Lighter colors indicate values approaching $1$ while darker colors indicate values approaching $0$.}}
    \label{fig:yolo_only_risk}
    \vspace{-4mm}
\end{figure}

First, we find the optimal design parameters for a baseline AEBS with only an object detector.  We evaluate the design with deadlines every $250$ ms, which corresponds to processing a video stream at $4$ Hz.  In Fig.~\ref{fig:yolo_only_risk} we plot $P(E^{base}_0)$ and $P(E^{base}_1)$ across the entire design space.  As we expected, when the input image size is small, $P(E^{base}_0)$ tends to be large as it is more difficult for YOLO to identify objects.  Likewise when image sizes are large, $P(E^{base}_0)$ is also high due to deadline misses.  $P(E^{base}_1)$ appears steady across the range of sizes indicating robustness to false positives.  While $P(E^{base}_0)$ is minimized with a lower threshold, $P(E^{base}_1)$ is minimized with the highest threshold.  By combining both events with risk, we are able to find a solution that compromises between the two metrics. 
Table~\ref{tab:design_yolo_only} (YOLO Only) shows the minimum risk and corresponding average utilization for the baseline system along with the optimal $\Lambda$ and $T$.  We calculate risk assuming $S(E^{base}_0)=3$ as this can lead to a head on collision and $S(E^{base}_1)=1$ as this event can cause the vehicle to be struck from the rear~\cite{krampe2020}.
\begin{table}
  \vspace{1mm}
  \centering
  \caption{\footnotesize{\textsc{Comparison of optimal design attributes for the baseline AEBS and combined YOLO-OOD AEBS at various sampling frequencies.}}}
  \begin{tabular}{|>{\centering\arraybackslash}p{0.12\linewidth}|>{\centering\arraybackslash}p{0.16\linewidth}|>{\centering\arraybackslash}p{0.15\linewidth}|>{\centering\arraybackslash}p{0.15\linewidth}|>{\centering\arraybackslash}p{0.15\linewidth}|}
    \hline
    \textbf{Config}.&YOLO Only @4Hz&Combined @3Hz&Combined @4Hz&Combined @5Hz\\
    \hline
    \textbf{Min. Risk}&$0.6337$&$0.2735$&$0.2104$&Infeas.\\
    \hline
    \textbf{YOLO Thresh.}&$0.20$&$0.61$&$0.14$&Infeas.\\
    \hline
    \textbf{YOLO Size.}&$192\times192$&$384\times384$&$160\times160$&Infeas.\\
    \hline
    \textbf{OOD Thresh.}&N/A&$0.99$&$0.99$&Infeas.\\
    \hline
    \textbf{OOD Size.}&N/A&$16\times16$&$64\times64$&Infeas.\\
    \hline
    \textbf{OOD Arch.}&N/A&$\beta$-VAE&$\beta$-VAE&Infeas.\\
    \hline
    $\bar{U}$&$0.9252$&$0.6646$&$0.9232$&Infeas.\\
    \hline
\end{tabular}
  \label{tab:design_yolo_only}
  \vspace{-5mm}
\end{table}

\subsection{Risk Minimization for AEBS with OOD Detector}
Given the minimum risk baseline in Table~\ref{tab:design_yolo_only}, we use our design strategy to select an EC / OOD detector combination that lowers risk below the baseline case while not exceeding the baseline's $\bar{U}$. We set the number of partitions to $4$ as described in Table~\ref{tab:part}, and set $n_{init}$ to $5$. When gathering timing data, we allow the OOD detector to always run first when a new input arrives.  This design choice ensures that the OOD detector gets a chance to execute as YOLO blocks the embedded GPU for nearly the entire available duration, even at lower input sizes.
\begin{table}
\vspace{-5mm}
  \centering
  \caption{\footnotesize{\textsc{Partitions used in our co-design.}}}
  \begin{tabular}{|>{\centering\arraybackslash}p{0.07\linewidth}|>{\centering\arraybackslash}p{0.12\linewidth}|>{\centering\arraybackslash}p{0.12\linewidth}|>{\centering\arraybackslash}p{0.45\linewidth}|}
    \hline
    \textbf{Part.}&\textbf{YOLO Sizes}&\textbf{OOD Sizes}&\textbf{OOD Arch.}\\
    \hline
    1&64--272&8--116&\{$\beta$-VAE, reconstruction based\}\\
    2&272--480&8--116&\{$\beta$-VAE, reconstruction based\}\\
    3&64--272&116--224&\{$\beta$-VAE, reconstruction based\}\\
    4&272--480&116--224&\{$\beta$-VAE, reconstruction based\}\\
    \hline
\end{tabular}
  \label{tab:part}
\end{table}
\begin{figure}
    \vspace{1.5mm}
    \centering
    \includegraphics[width=0.47\textwidth]{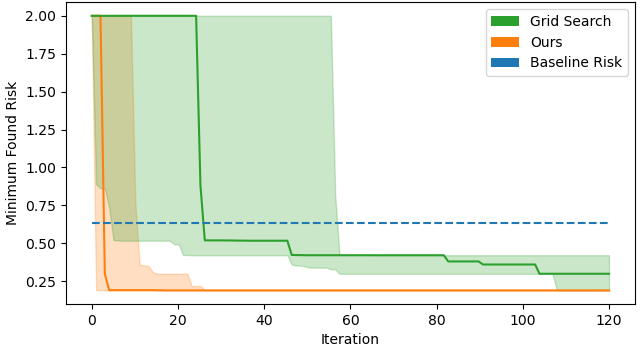}
    \caption{\footnotesize{Risk versus number of iterations for our Bayesian optimizer vs. grid search. The solid line corresponds to the median lowest risk at that epoch and the shaded region shows the $75^{th}$ to $25^{th}$ percentile evaluated over $100$ trials with different random seeds.}}
    \label{fig:opt}
    \vspace{-6mm}
\end{figure}
Fig.~\ref{fig:opt} shows the convergence of our Bayesian optimization strategy (orange) compared with grid search (green). The grid search was conducted by picking a random point in the search space and sweeping all parameters (increments of 8 px. for OOD detector size and 32 px. for YOLO input size).  Our modified Bayesian optimization converges faster than grid search, finding the minimum solution within $30$ iteration $75\%$ of the time.

\begin{figure}
    \vspace{1.5mm}
    \centering
    \includegraphics[width=0.47\textwidth]{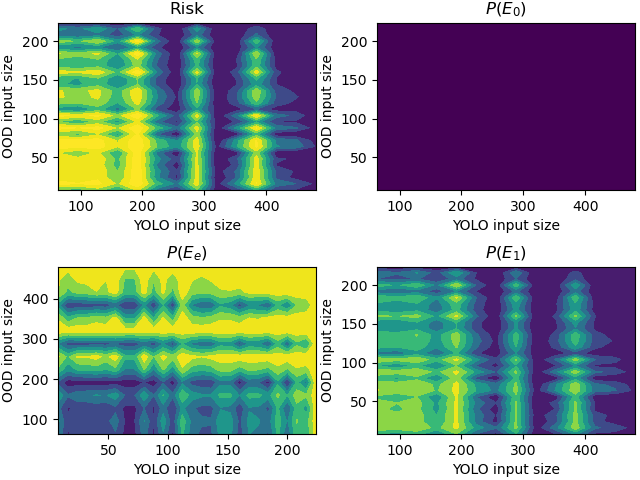}
    \caption{\footnotesize{Projection of risk, $P(E_e)$, $P(E_0)$, and $P(E_1)$ onto OOD and YOLO design parameters.  All parameters not specified in a plot are set to the values that produce the minimum risk. Upper left: impact of input sizes on risk.  Lower left: impact input sizes on $P(E_e)$. Upper right: impact of input sizes on $P(E_0)$.  Lower right, impact of input sizes on $P(E_1)$.}}
    \label{fig:risk_ood_first}
    \vspace{-5mm}
\end{figure}

The minimum risk and corresponding configuration are shown in Table~\ref{tab:design_yolo_only} (Combined @4Hz).  We note that in all experiments the $\beta$-VAE OOD detector outperformed the reconstruction based OOD detector in terms of risk due to the reconstruction based OOD detector's high probability of deadline misses.  We also note that the YOLO parameters for the minimum risk configuration are different than baseline case, indicating the importance of co-design.  Fig.~\ref{fig:risk_ood_first} shows a visualization of the risk surface for the design of YOLO based AEBS with OOD detector at input frequency $4$ Hz. We observe that at the selected thresholds, small input images for the OOD detector and YOLO tend to dramatically increase risk (Fig.~\ref{fig:risk_ood_first}, upper left).  This makes sense as in this minimum risk configuration, a majority of the remaining risk is supplied by $P(E_1)$ (Fig.~\ref{fig:risk_ood_first}, bottom right).  As the severity of $P(E_0)$ causes it to contribute more to the overall risk, a configuration was selected during the minimization where $P(E_0)$ is low compared to $P(E_1)$ and relatively invariant with respect to other design parameters (Fig.~\ref{fig:risk_ood_first}, upper right).  In Fig.~\ref{fig:risk_ood_first} (bottom left) we also see $P(E_e)$, the probability of YOLO missing a deadline, across design parameters.  As expected, we see that this increases for larger input sizes, but note that it is not the dominant factor in our risk plot.  This makes sense as deadline misses contribute to $E_0$, which is small at the optimal solution compared to the contribution of $E_1$, where more deadline misses can reduce the FPR.

Finally, we perform the risk minimization again, but assume a deadline of $200$ ms ($5$ Hz) to see if our design methodology will allow us to achieve a less risky solution that can sustain a higher sampling rate and is therefore applicable at higher vehicle speeds than the baseline (Table~\ref{tab:design_yolo_only}: Combined @5Hz).  Unfortunately there are no feasible solutions at this rate given our embedded hardware.  We also reran the optimization with deadlines every $333$ ms ($3$ Hz) to see if the extra execution time would allow us to further reduce risk at the expense of vehicle speed (Table~\ref{tab:design_yolo_only}: Combined @3Hz).  In this case average utilization is decreased well below the baseline due to the extra available time, but the overall minimum risk is on par with the 4 Hz case.  In this case the functional performance of the OOD detector and YOLO object detector limits the overall risk.  This indicates that while meeting deadlines is critical to finding a minimal risk solution, setting arbitrarily long deadlines (even if vehicle's speed is slow enough to allow it) does not necessarily help reduce risk further.

\section{LIMITATIONS}
While this work shows the utility of an OOD detector as a safety monitor and the advantages of using risk in the co-design of LECs, there are still challenges that must be addressed to bridge the gap between a simulated case study and real transportation systems.  Most importantly, there is an implicit assumption that the datasets used for training and validation incorporate the same distribution of edge cases that the system will experience during operation.  If this assumption is not met, the risk returned by the co-design is not valid and guaranteeing that this assumption is met may not be feasible for real-world datasets.  In this paper a simulated dataset was used to help ensure sufficient coverage of scenarios and reduce the time required to collect data.  However, for a real dataset, collecting edge case scenarios and OOD samples could prove dangerous or costly and simulation of such scenarios does not guarantee that the analysis is valid for the corresponding physical system.

Furthermore, we assume the results of the OOD detector and EC are independent across time when determining the probabilities used in the risk analysis.  In a practical system, such an assumption is not feasible as previous control actions affect future samples.  Also, environmental conditions that determine if a sample is OOD are unlikely to change much between consecutive samples.  Incorporating these effects into the risk analysis is a future area of research.


Additionally, this work did not include a study of multimodal AEBSs.  Multimodal sensor data is common in robotic and transportation systems and, in principle, such a system can still be modeled as a binary classifier or ensemble of binary classifiers.  However, given a specific system architecture, multimodal input data may allow additional architectural enhancements that can help reduce risk.

\section{CONCLUSION}
We addressed the problem of co-designing an OOD detector and LEC for use in an AEBS.  While previous works have only focused on reducing the execution times of individual components or increasing their accuracy, our experiments show that the tradeoff between the functional and non-functional performance of each component needs to be taken into consideration when designing for safety. We observed that with our design methodology we were able to reduce risk below that of a baseline system while maintaining the same resource utilization, but that a design approach where both components are developed independently may not yield such a solution.  We also demonstrated that our design methodology reduces the time to find a minimal risk solution.  This work shows that deploying OOD detectors as safety monitors is feasible, but must be done as part of a co-design process to prevent inadvertently increasing risk.


\end{document}